\documentclass[runningheads]{llncs}
\usepackage[T1]{fontenc}
\usepackage{hyperref}
\usepackage{graphicx}
\usepackage{color}

\usepackage[normalem]{ulem}
\usepackage{amsmath,amssymb,amsfonts}
\usepackage{multirow}
\usepackage{makecell}
\usepackage{tablefootnote}
\usepackage{subfig}
\usepackage{textcomp}
\usepackage{xcolor}
\usepackage{tikz}
\usetikzlibrary{bayesnet}
\makeatletter
\newcommand*{\inlineequation}[2][]{%
	\begingroup
	\refstepcounter{equation}%
	\ifx\\#1\\%
	\else
	\label{#1}%
	\fi
	\relpenalty=10000 %
	\binoppenalty=10000 %
	\ensuremath{%
		#2%
	}%
	~\@eqnnum
	\endgroup
}
\makeatother
\begin{document}
\title{Material Microstructure Design Using VAE-Regression with Multimodal Prior}
\titlerunning{Microstructure Design Using VAE-Regression}
\author{Avadhut Sardeshmukh\inst{1,2}\orcidID{0000-0002-2374-4083}$^*$ \and
Sreedhar Reddy\inst{1} \and
BP Gautham\inst{1}\orcidID{0000-0002-1234-8632} \and
Pushpak Bhattacharyya\inst{2}}
\authorrunning{A. Sardeshmukh et al.}
\institute{TCS Research, Tata Consultancy Services Ltd, Pune, 411057, India \and
Department of Computer Science and Engineering, Indian Institute of
Technology Bombay, Mumbai, 400076, India\\
$^*$Corresponding Author.
\email{avadhut.sardeshmukh@tcs.com,\\ \{sreedhar.readdy,bp.gautham\}@tcs.com, pb@cse.iitb.ac.in}\\
}

\maketitle              % typeset the header of the contribution
\begin{abstract}
	We propose a variational autoencoder (VAE)-based model for building forward and inverse structure-property linkages, a problem of paramount importance in computational materials science. Our model systematically combines VAE with regression, linking the two models through a two-level prior conditioned on the regression variables. The regression loss is optimized jointly with the reconstruction loss of the variational autoencoder, learning microstructure features relevant for property prediction and reconstruction. The resultant model can be used for both forward and inverse prediction i.e., for predicting the properties of a given microstructure as well as for predicting the microstructure required to obtain given properties. Since the inverse problem is ill-posed (one-to-many), we derive the objective function using a multi-modal Gaussian mixture prior enabling the model to infer multiple microstructures for a target set of properties. We show that for forward prediction, our model is as accurate as state-of-the-art forward-only models. Additionally, our method enables direct inverse inference. We show that the microstructures inferred using our model achieve desired properties reasonably accurately, avoiding the need for expensive optimization loops.
\end{abstract}
\section{Introduction}\label{sec:intro}
Materials science and engineering involve studying different materials, their processing and the resulting properties that govern the performance of the material in operation. Processes such as heating, tempering and rolling modify the material's internal structure, altering properties such as tensile strength, ductility and so on. The structure is commonly represented by microscopy images known as the microstructure, which contain information about the micro-constituents (also known as phases), the grains, their geometry (shape, size), and their orientations. These structural features impact the material properties. Modeling the relationships between processing, structure and properties (also known as the P-S-P linkages) is at the core of computational materials science. Materials scientists and engineers are often interested in inverse analysis i.e., predicting the candidate structure for target properties and the processing route required to get the target structure. This involves systematically exploring a large design space consisting of several possible initial compositions, processing steps, parameters of these processing steps and the resulting structures. Also, the problem is often ill-posed since multiple processing routes can lead to the same structure and multiple structures can lead to the same selected target properties.  Traditionally, materials scientists have used a combination of experimentation and physics-based numerical simulations for inverse analysis. However, experimental exploration has limitations because of the time and cost involved. On the other hand, physics-based models, which are based on solving underlying differential equations, are only useful for predicting the forward path. That is, predicting structure from composition and processing conditions, and properties from structure. They cannot be used directly for the inverse problem. Instead, they have to be used inside an optimization loop~\cite{ankit15PropStruct}. However, physics-based models are often computationally too expensive to be useful for design space exploration and optimization. 

Machine learning can be an alternative to physics-based simulations for forward prediction. For example, deep convolutional neural networks have been used for predicting properties from microstructure images (\cite{kalidindi3dcnn,listructprop}). Recently, probabilistic deep generative models have been proposed to learn features from unlabeled (i.e. no properties data) microstructure images and then train a property-prediction model in the low-dimensional feature space using small labeled data (\cite{cangVAE,texturevae}). This is an advantage since labeled data is more challenging to get. However, these models are not capable of inverse inference themselves. They still have to be used inside an optimization loop. While probabilistic deep generative models can be leveraged for direct inverse inference, they have not been explored much for this purpose. Our work is aimed at addressing this gap.

We propose a probabilistic generative model of structure-property linkage by combining variational autoencoder (VAE) with regression such that the joint model can be used for both forward and inverse inference without requiring the optimization step for the inverse.  
The VAE and the regression model are joined through the VAE prior by making the prior conditional on the predicted property. After training, latent representations for a target property value can be sampled from the conditional prior and decoded to get microstructures with that property value. Further, since there can be more than one microstructure for a target property value, the commonly used uni-modal Gaussian prior does not model this accurately. So we replace it with a mixture of Gaussians.  

Our contributions are - i) a method for forward prediction of properties from structure, ii) a method for direct inverse prediction of candidate structures given the target properties, effectively handling ill-posedness. Using a reference dataset of 3-D microstructures and elasticity properties, we show that our model is as accurate as state-of-the-art methods for forward inference (table~\ref{tab:forwardaccuracy}) while additionally enabling direct inverse inference. For inverse inference, we show that the microstructures inferred using our model achieve the desired properties reasonably accurately (fig.~\ref{fig:invinfbarchart}). Further, we show that optimal points lie very close to these inferred microstructures and more precise solutions can be quickly found by searching in the neighborhood of these microstructures using a detailed physics based model (fig.~\ref{fig:optpca} and~\ref{fig:invinfoptbarchart}). 
\section{Methodology}\label{sec:methodology}
Variational autoencoders (\cite{vae_kingma}) pose the problem of representation learning as probabilistic inference with the underlying generative model $p(x,z) = p(x|z)p(z)$, where $x$ is the input and $z$ is the latent representation. The posterior $p(z|x)$ is to be inferred. An approximate posterior $q(z|x)$ is found by minimizing the Evidence Lower BOound (ELBO):
\inlineequation[eqn:elbo]{\mathcal{L} = -D_{KL}(q(z|x)\|p(z)) + \mathbb{E}_{q(z|x)}[log ~p(x|z)]}

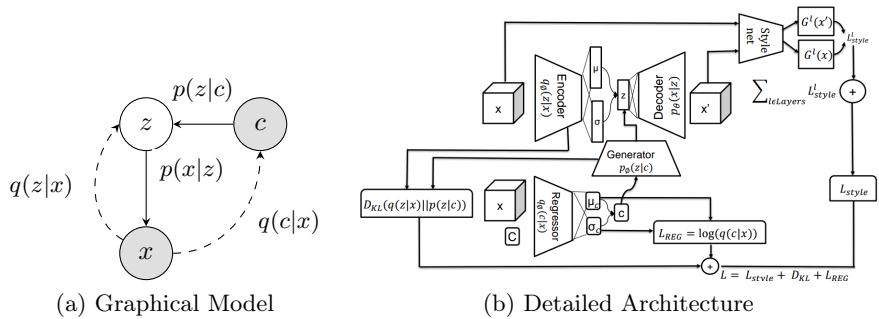
\begin{figure}[!t]
	\centering
		\subfloat[Graphical Model]{
			\begin{tikzpicture}
			% Define nodes
			\node[obs]                  (x) {$x$};
			\node[latent, above=of x] (z) {$z$};
			\node[obs, right=0.8cm of z]  (c) {$c$};
			% Connect the nodes - generative model
			\draw[-stealth] (c) -- (z) node [midway, above=5pt, fill=none]{$p(z|c)$} ; %
			\draw[-stealth] (z) -- (x) node [near start, right=2pt, fill=white]{$p(x|z)$} ; %
			% Connect the nodes - variational model
			\draw[dashed,-stealth] (x) to [out=150,in=230] node [left=5pt, fill=white]{$q(z|x)$} (z.west) ;
			\draw[dashed,-stealth] (x.east) to [out=0,in=270] node [right=5pt, fill=white]{$q(c|x)$} (c.south) ;		
			\end{tikzpicture}
			\label{fig:vae_archi_a}}
		\hfil
		\subfloat[Detailed Architecture]{
		\includegraphics[scale=0.18, clip, trim=0 0.25cm 0 0.55cm]{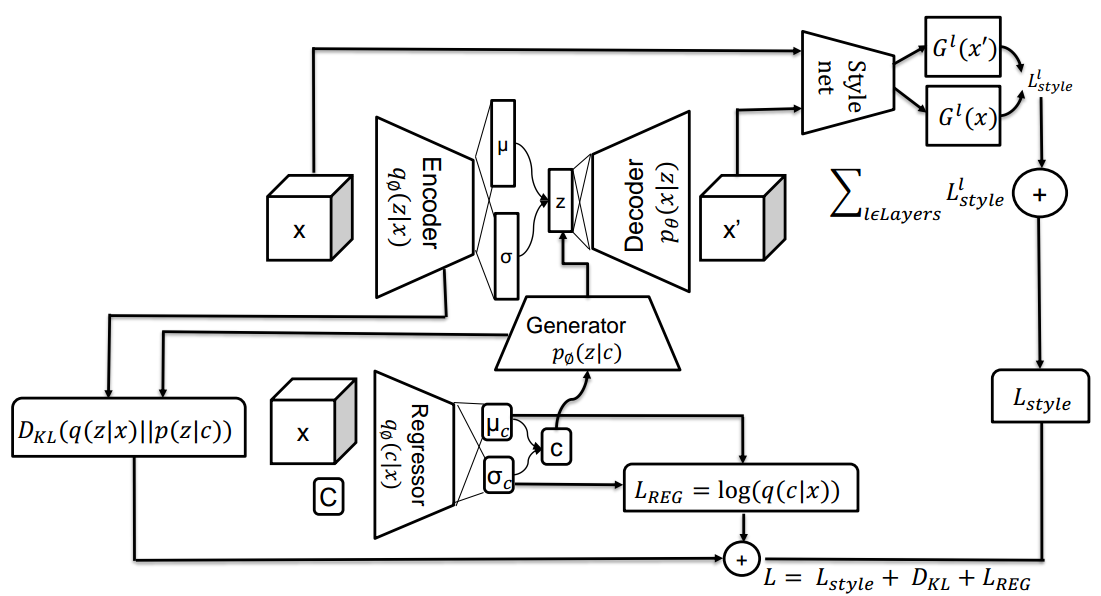}
		\label{fig:vae_archi_b}}
	\caption{Architecture : VAE-Regression with Style Loss.} 
	\label{fig:vae_archi}
\end{figure}

The KL-Divergence term enforces the prior $p(z)$ as a regularization while the second term quantifies how well an $x$ is reconstructed.\\ 
%\label{sec:methodologyvaereg}
\textbf{VAE-Regression}
The use of VAE in semi-supervised or supervised regression settings (predicting a scalar or vector of real numbers from an image) has been relatively less explored.  
A ``VAE for regression'' model was proposed by Zhao et al.~\cite{vae_for_regression} for predicting a subject's age from their structural Magnetic Resonance (MR) images, as follows. Assuming that the latent representation $z$ is also dependent on the quantity $c$ to be predicted, the generative model is: $p(x,z,c) = p(c)p(z|c)p(x|z)$, leading to a two-level prior (see fig.~\ref{fig:vae_archi_a}). %$p(c)$ is the prior on $c$ whereas $p(z|c)$ is the prior on latent representation, conditioned on age. 
The approximate posterior $q(z,c|x)$ is found using variational inference assuming that $q$ factorizes as $q(z,c|x) = q(z|x)q(c|x)$~\footnote{The mean-field assumption~\cite{vi_davidblei}, commonly used in variational inference derivations (e.g.,~\cite{cpvae_expanded})}. Note however that the dependence between $z$ and $c$ is indirectly preserved through the prior $p(z|c)$. With that assumption, Zhao et al.~\cite{vae_for_regression} derive the modified ELBO as:
\begin{equation}
	\mathcal{L} = \underbrace{-D_{KL}\left(q(c|x)\|p(c)\right)}_{\text{Regression loss}} + 
	\underbrace{\mathbb{E}_{q(z|x)}[log ~p(x|z)]}_{\text{Rec loss}} - \underbrace{\mathbb{E}_{q(c|x)}[D_{KL}(q(z|x)\|p(z|c))]}_{\text{Regularization (cond. prior)}}
	\label{eqn:elbovaereg}
\end{equation}
In supervised settings, the first term can be replaced by $log ~q(c|x)$, which is proportional to the mean squared error (i.e. regression loss) when $q(c|x)$ is a Gaussian. So $q(c|x)$ is parameterized by a ``regressor'' network. The second term -- the reconstruction loss -- is the same as in the original ELBO from equation \eqref{eqn:elbo}, so $q(z|x)$ and $p(x|z)$ are parameterized by ``encoder'' and ``decoder'' networks, respectively. The last term is a counterpart of the regularizer from equation \eqref{eqn:elbo}, except the prior is now conditional on $c$ and the KL divergence is now in expectation with respect to $q(c|x)$. %. This is the term that connects the regressor and the VAE. The expectation is  
This term encourages the encoder posterior $q(z|x)$ and the conditional prior $p(z|c)$ to be similar, aligning the features for reconstruction and property prediction. Since the expectation is with respect to $q(c|x)$, it links the VAE and the regressor. During training, the expectation is estimated Monte-Carlo, using the $c$ predicted by the regressor (i.e., one sample from $q(c|x)$). The distribution $p(z|c)$ is parameterized by a ``generator'' network. After training, inverse inference can be performed by sampling latent representations for a target $c$ from the generator and decoding them.

We make two modifications to the original formulation from \cite{vae_for_regression} to make it suitable to our use case. First, we replace the reconstruction loss with the style loss, which is based on comparing the statistics of the microstructures. Second, instead of the standard, uni-modal Gaussian prior, we incorporate a Gaussian mixture prior which models the many-to-one structure-property relation better.\\
\textbf{VAE-Regression with Style Loss}
The reconstruction loss from vanilla VAE objective function typically leads to a pixel-by-pixel comparison between input and reconstruction, which is not suitable for microstructure images~\cite{texturevae}.  
Rather, only a comparison between statistics is suited to quantify the difference between two microstructures.
The ``style loss'', originally proposed for the problem of style-transfer~\cite{gatys}, is based on a comparison between statistics. It is the sum of squared differences between Gram matrices of the input and reconstruction as computed from a deep pre-trained network such as VGG19~\cite{vgg19}. 
Say layer $l$ has $C_l$ feature maps of size  $W_l \times H_l$ then the Gram matrix at layer $l$ is $G^l_{ij} = \Sigma_k F^l_{ik} F^l_{jk}$, and the style loss is:
$
	\mathcal{L}_{style}(x,\hat{x}) = \sum_{l=0}^{L} w_l \left[ \frac{1}{4C_l^2 M_l^2} \sum_{i,j} (G^l_{ij} - \hat{G}^l_{ij})^2 \right]
	\label{eqn:styleloss}
$
Where, $M_l = W_l*H_l$ and $F^l$ is the $C_l \times M_l$ matrix of flattened feature maps. We propose to replace the reconstruction loss from the modified ELBO (equation~\eqref{eqn:elbovaereg}) with the style loss. The architecture of ``VAE-regression'' after incorporating the style loss is shown in Fig.~\ref{fig:vae_archi_b}.\\
\textbf{Multi-modal Prior} 
Since multiple structures can lead to the same properties, there could be more than one likely latent representation $z$, for a given $c$. So, the standard Gaussian prior, which a uni-modal distribution is not suitable for modeling $p(z|c)$. Instead, we propose a mixture-of-Gaussians prior
$p(z|c) \sim \sum_{k=1}^{K} \pi_k N_k(\mu_k, \sigma_k^2)$ with $K$ components, where $\pi$ are the probabilities of components. We assume a diagonal co-variance matrix for all mixture components. 
The generator network outputs $K$ pairs of $\mu$ and $\sigma$ and the $K$ component probabilities. Note that $K$ is a hyperparameter that needs to be tuned.
Since the posterior $q(z|x)$ is still a Gaussian, the reparameterization trick from the original VAE works. The only difficulty is in computing the KL-Divergence of the posterior from the conditional prior, that is the third term from equation~\eqref{eqn:elbovaereg}. We need to compute the KL-Divergence of a Gaussian from a mixture-of-Gaussians, which is intractable. We use a variational approximation proposed in speech recognition literature~\cite{kld_gmm1}, which for Gaussian mixtures $f$ and $g$ is:
$
D_{KL}(f\|g) \approx \sum_k \boldsymbol{\omega}_k log \dfrac{\sum_{i} \boldsymbol{\omega}_{i}e^{-D_{KL}(f_k\|f_i)}}{\sum_{j}\boldsymbol{\pi}_j e^{-D_{KL}(f_k\|g_j)}}
\label{eqn:gmmkld}
$
where $f_k$, $f_{i}$ and $g_j$ denote the component Gaussians of $f$ and $g$ and $\boldsymbol{\omega}$ and $\boldsymbol{\pi}$ are their component weights respectively. Since in our case the first distribution $f$ which is the posterior,  is a uni-modal Gaussian, the numerator reduces to 1 (since $D_{KL}(f\|f) = 0$). So the expression is:
\begin{equation}
D_{KL}(f\|g) \approx log \dfrac{1}{\sum_{j}\boldsymbol{\pi}_j e^{-D_{KL}(f\|g_j)}}
\label{eqn:gmm_gauss_kld}
\end{equation}
We replace the third term from equation \eqref{eqn:elbovaereg} with the RHS of eq.~\eqref{eqn:gmm_gauss_kld} to get our final loss function as follows:
%\begin{small}
\begin{equation}
	\mathcal{L}_{VAE-REG} = -log\left(q(c|x)\right) + 
	\mathcal{L}_{style} - log \dfrac{1}{\sum_{j}\boldsymbol{\pi}_j e^{-D_{KL}(q(z|x)\|p_j(z|c))}}
	\label{eqn:final_loss_function}
\end{equation}
\section{Related Work}
\label{sec:relatedwork}
Due to recent advancements in machine learning, there is a renewed interest among materials scientists to leverage state-of-the art deep learning models for P-S-P linkages~\cite{kalidindi3dcnn,listructprop,kalidindi3dcnn2}. However, all
these works focus on forward prediction using discriminative models. While Cang et al.~\cite{cangVAE} propose a generative VAE model, new images are generated unconditionally and used as additional training data for a downstream forward property-prediction model. In contrast, our approach enables forward and direct inverse inference combining VAE and regression through a conditional prior.

Recently, deep generative models such as variational autoencoders and Generative Adversarial Networks (GAN) have been explored for inverse inference in structure-property linkage~\cite{vaegpr,ankit15PropStruct} and other similar problems such as drug design~\cite{vae_opt_prob1,vae_opt_acs}. These methods train a forward regression model from the GAN/VAE latent space to properties. For inverse inference, an optimization loop is setup around the forward model. While the GAN/VAE latent space enables efficient navigation through the space of microstructures, it does not alleviate the need for optimization. As opposed to this, in our approach once the model is trained, inverse inference is same as prediction and does not need optimization. More recently, Mao et al.~\cite{ankit22PropStruct} extended the GAN-based inverse inference method from Yang et al.~\cite{ankit15PropStruct} using mixture density networks (MDN). The GAN is trained on microstructure images and used to create training data for MDN. A set of images are generated from the GAN and the corresponding properties are obtained through FEM simulations on these images. The MDN is then trained to predict the GAN inputs $z$ from the properties $c$. While this method does not need optimization every time, it can't utilize existing labeled data between microstructures and properties since the MDN needs pairs of $z$ and $c$.

In machine learning literature, use of VAE in semi-supervised settings has been explored for various objectives such as conditional generation~\cite{cvae_m2}, learning disentangled representations~\cite{ccvae}, multi-modal representation learning~\cite{meme} and so on. Most works on conditional generation treat the label ($c$) as one more latent variable, leading to a graphical model like $z\rightarrow x \leftarrow c$, where $x$ is the input and $z$ is the latent representation. However in our problem, $c$ represents the material property which affects the microstructure features relevant for that property. So we treat the label as an auxiliary variable affecting the latent representation, leading to $c \rightarrow z \rightarrow x$.  
Probably the closest to our work is Characteristic Capturing VAE \cite{ccvae}, which focuses on capturing the label characteristics in the latent representation. While our probabilistic generative model is same as theirs, the variational assumptions are slightly different. We assume that the variational posterior factorizes as $q(c,z|x) = q(c|x)q(z|x)$ whereas they assume $q(c,z|x) = q(c|z)q(z|x)$. Our assumption greatly simplifies the mathematical derivation while preserving the essential dependencies indirectly (via KL-divergence). 

Some recent works have focused on incorporating stronger priors (including Gaussian mixture) in VAE (e.g.,~\cite{gmvae,cpvae_expanded}) to improve the quality of generations. However, in these formulations, $c$ is typically a discrete latent variable that represents hidden modalities of data, making the joint $p(z,c)$ a Gaussian mixture. Whereas our motivation for a multi-modal prior comes from the many-to-one relationship between structures $x$ and properties $c$. So $c$ is observed and continuous, and the conditional $p(z|c)$ itself is a Gaussian mixture. This introduces an intractable KL divergence term in our loss function unlike others, which we have dealt with using variational approximation, as explained in section~\ref{sec:methodology}.
\section{Experimental Results}\label{sec:experiments}
\begin{figure}
	\centering	
	\subfloat[{$\sigma=[1,7,7]$}]{
		\includegraphics[scale=0.1]{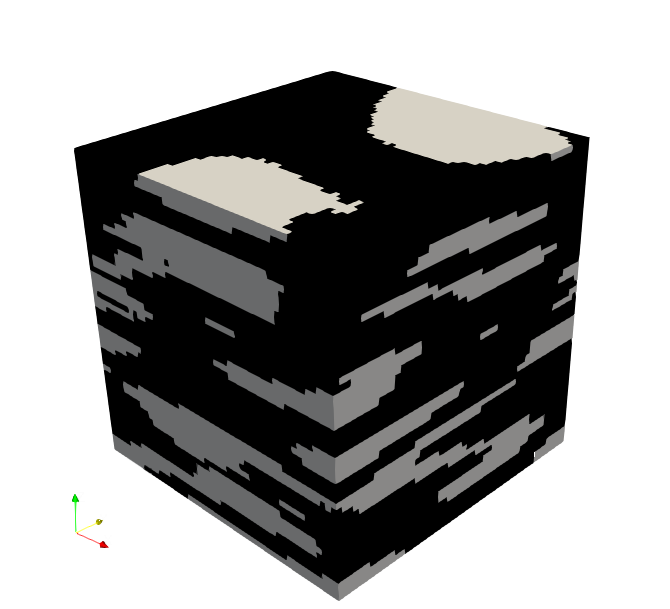}
		\label{fig:microswithc25a}}
	\hfil
	\subfloat[{$\sigma=[7,1,1]$}]{
		\includegraphics[scale=0.1]{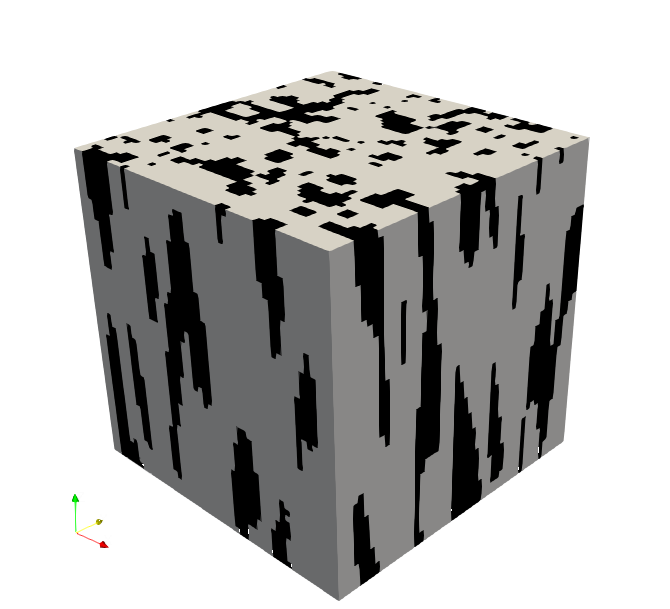}
		\label{fig:microswithc25b}}
	\hfil
	\subfloat[$C_{11}$ vs. volume fraction of black phase]{
	\includegraphics[scale=0.22, clip, trim=0cm 0cm 1cm 1.5cm]{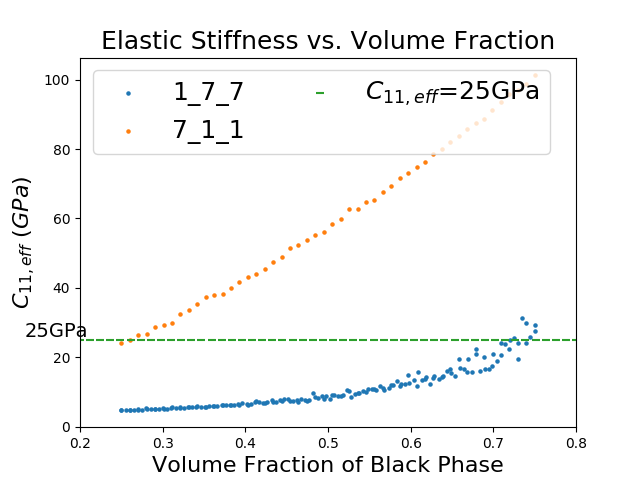}
	\label{fig:microswithc25c}}
	\caption{Two microstructures with $C_{11} \approxeq 25GPa$. The volume fraction of black phase in (a) is 74.49\% and in (b) is 26.02\%, as shown by the green line in (c)}
	\label{fig:microswithc25}
\end{figure}

We now describe the results obtained on a dataset of 3-D microstructures of a high-contrast composite and the associated elastic stiffness property.\\
\textbf{Dataset} We use a dataset presented in Fernandez-Zelaia et al.~\cite{kalidindidatasource} as part of their work on an efficient finite element method for micro-mechanics simulation in high-contrast composites. The authors first synthetically generated a large ensemble of voxelized 3D microstructures with diverse morphological features. This was done by starting with random 3D inputs of size 51x51x51 in [0,1] and convolution with various Gaussian filters with zero mean and diagonal covariance, and thresholding the results to obtain a binary microstructure. The diagonal of the covariance matrix of the 3D Gaussian filters was of the form $\sigma = [i,j,k], i,j,k \in \{1,3,5,7\}$. Thus there are 64 different Gaussian filters, which were applied to 150 random inputs resulting in $\sim$8900 3D microstructures with a wide variety of morphologies.
For example, $\sigma = [1,1,1]$ results in small, isotropic grains whereas $\sigma= [7,7,7]$ results in very large grains. Other asymmetric choices such as $\sigma = [7,1,1]$ or $[1,1,7]$ result in anisotropic grains with elongation in the corresponding directions. Please see Fig.~\ref{fig:microswithc25a} and \ref{fig:microswithc25b} for examples.

\begin{figure}[!t]	
	\begin{minipage}{0.45\textwidth}
		\centering
		\subfloat[Target $C_{11} = 30GPa$]{
		\includegraphics[scale=0.22]{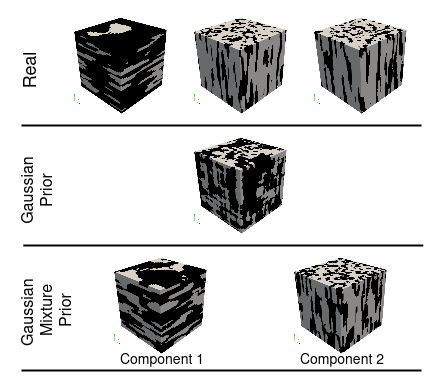}
		\label{fig:invinfreconb}}	
	\end{minipage}
	\hfil
	\begin{minipage}{0.45\textwidth}
		\centering
		\subfloat[Inverse Inference - Abs. \% Error]{
			\includegraphics[scale=0.25, clip, trim=0cm 1.0cm 0cm 0.8cm]{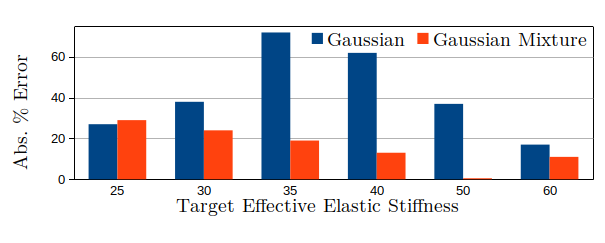}
			\label{fig:invinfbarchart}}		
	\end{minipage}
	\caption{(a) Inverse inference for target $C_{11} = 30GPa$. Top row shows real microstructures with the target $C_{11}$. The microstructures inferred using a uni-modal Gaussian prior (middle row) tend to be like an average, whereas the multi-modal Gaussian mixture prior learns the multiple possible solutions under separate mixture components (last row). (b) Evaluation of inverse inference through FEM simulations to get the achieved $C_{11}$ and compute the absolute \% error between target and achieved $C_{11}$. The multi-modal prior learns multiple solutions under separate components unlike the uni-modal Gaussian prior, leading to more accurate inference (error within 20\% in most cases).}
	\label{fig:invinfrecon}
\end{figure}

The authors further estimate the effective elastic stiffness property of these microstructures using FEM simulations~\cite{kalidindidatasource}. 
The crucial assumptions are: The black and white colors correspond to the hard and soft phases, with Young's moduli 120GPa and 2.4GPa, respectively, and the Poisson's ratio for both phases is 0.3.
Elastic stiffness relates stress $\sigma$ with strain $\epsilon$. Since these are second-rank tensors, a fourth-rank tensor is required to relate them, i.e.,
$\sigma_{ij} = C_{ijkl} \epsilon_{kl}$ (Generalized Hooke's law \cite{physicsbook}).
The effective elastic stiffness parameter mentioned above is the element $\langle1,1,1,1\rangle$ of the stiffness tensor $C$, and is denoted as $C_{11}$ for short.\\ 
\textbf{Training} We split the data into 60\% training, 20\% validation, and 20\% test data. The validation data is used only to decide when to stop training (we stop when the validation loss does not decrease for ten consecutive epochs).  
The model is trained end-to-end using Adam optimizer with learning rate 0.0005, $\beta_1 = 0.75$, $\beta_2 = 0.999$ and batch size 8. We found that with 16 latent dimensions and a weight of 10 for regression loss (after normalizing the scales of all three losses) we got the best forward prediction accuracy with good reconstructions. More details of architecture, training and hyperparameter tuning  appear in appendix~\ref{sec:appA}.\\ 
\textbf{Forward Inference} Table~\ref{tab:forwardaccuracy} shows accuracy in the prediction of $C_{11}$ using different methods. MAPE is the mean absolute percentage error ($MAPE=1/n\Sigma_i|y_i-\hat{y}_i|/\bar{y}$, where $\bar{y}$ is the mean observed value and $\hat{y}$ are the predictions) and $R^2$ is the coefficient of determination. The first block corresponds to two physics-based methods, the numbers reproduced from~\cite{kalidindi3dcnn2}. The next block corresponds to VAE and Gaussian process regression (GPR) trained separately. Appendix~\ref{sec:appB} shows similar results obtained using other regression models such as support vector regression.
The second last row corresponds to a state-of-the-art 3D CNN~\cite{kalidindi3dcnn2} for forward prediction only (we implemented this to reproduce the results), while the last row corresponds to our model. These numbers were obtained by averaging over ten random initializations and train-test splits (standard deviations in brackets). While our model is much more accurate than separately trained VAE and regression, it is comparable to the state-of-the-art regression-only model (considering the standard deviations) and additionally provides direct inverse inference.\\
\begin{table}
	\centering
	\caption{Forward Prediction Accuracy. Our method performs much better than traditional methods (first block) and separately trained VAE and regression (second block), and is as accurate as the state-of-the-art fwd model (Reg-Only).}	
	\begin{small}
		\begin{tabular}{l|c|c}		
			\hline
			Method & MAPE & R$^2$  \\
			\hline
			V-R-H Avg~\cite{kalidindi3dcnn2}  & 46.66 & -\\
			2pt Stats+Reg~\cite{kalidindi3dcnn2} & 6.79 & -\\
			\hline
			VAE + GPR & 7.23 ($\pm$0.15) & 0.97 ($\pm$1e-3)\\		
			\hline		
			Reg-Only~\cite{kalidindi3dcnn2}
			& 3.69 ($\pm$0.17) & 0.99 ($\pm$4e-4) \\
			VAE-Reg (ours) & \textbf{3.50} ($\pm$0.24) & 0.99 ($\pm$5e-4)\\
			\hline		
		\end{tabular}
	\end{small}
	\label{tab:forwardaccuracy}
\end{table}
\noindent\textbf{Inverse Inference} The effective elastic stiffness property $C_{11}$ is largely dependent on the volume fractions. However, this function changes with the morphology. This is apparent from Fig.~\ref{fig:microswithc25}. Fig.~\ref{fig:microswithc25a} and ~\ref{fig:microswithc25b} show microstructures with $C_{11} \approxeq 25GPa$, generated from $\sigma=[1,7,7]$ and $\sigma=[7,1,1]$ having black phase volumes 74.49\% and 26.02\%, respectively.
Fig.~\ref{fig:microswithc25c}, shows $C_{11}$ against the volume fraction of the black phase in the microstructures from these two morphologies. Thus, a given value of $C_{11}$ can be achieved by multiple microstructures, possibly each coming from a different morphology with a different volume fraction of the black phase. This motivates the use of a multi-modal conditional prior.\\ %$p(z|c)$, so that there are multiple expected latent vectors $z$, for a target $c$. 
\hspace{\parindent}To demonstrate the effectiveness of multi-modal inverse inference, we chose a subset of microstructures from the data, corresponding to $\sigma=[1,1,7]$ and $\sigma=[7,1,1]$, which have remarkably different morphologies (see Fig.~\ref{fig:microswithc25a} and~\ref{fig:microswithc25b} for an example) and are well-separated in the property space, but with some overlap at the extremes, as shown in Fig.~\ref{fig:microswithc25c}. We perform inverse inference for six target values of $C_{11}$ spread across the complete range observed in the data. For each target $C_{11}$ value, we obtain the mean of the conditional prior (for the multi-modal prior, the means of each component) and decode it.
For illustration, we discuss the inverse inference for $C_{11} = 30GPa$. Fig.~\ref{fig:invinfreconb} shows the real (top) and inferred microstructures from the uni-modal Gaussian prior (middle) and the two components of the Gaussian mixture prior. Note that the uni-modal prior tends to infer an average of the two possible solutions. 
Whereas the Gaussian mixture prior learns the solutions separately under different mixture components, with suitable weights.\\
\hspace{\parindent}Further, we validated the inferred microstructures through FEM simulations to estimate the achieved properties. Fig.~\ref{fig:invinfbarchart} shows the absolute percentage error between target and achieved properties using the mean solutions under uni-modal and multi-modal priors for a range of $C_{11}$ values. The solutions inferred from the Gaussian mixture prior achieve the target properties better than those inferred from the uni-modal Gaussian prior. The difference is more pronounced for target $C_{11}$ values in the middle (e.g., 35GPa), where both the morphologies are likely. The average absolute error using Gaussian mixture prior is about 16\%, with an $R^2$ value of 0.97.\\
\textbf{Comparison with optimization}\\
As discussed in sections~\ref{sec:intro} and \ref{sec:relatedwork}, some recent works have proposed optimization in the latent space of VAE or GAN for inverse inference. However, exploring the latent space efficiently and finding multiple optima can be difficult~\cite{vae_opt_prob1}. Searching from multiple random initial points may be required to find good solutions. In contrast, our inverse inference method directly provides some good initial candidates in the optimal regions. Detailed search can then be efficiently performed in the neighborhood to reach high-quality solutions. 

\begin{figure}
	\begin{minipage}{0.47\textwidth}
	\centering
	\subfloat[{}]{
		\includegraphics[scale=0.25]{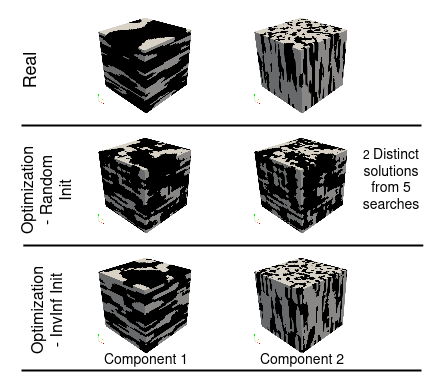}
		\label{fig:invinfopt}}
	\end{minipage}
	\hfil
	\begin{minipage}{0.47\textwidth}
		\centering
	\subfloat[{}]{
	\includegraphics[scale=0.17]{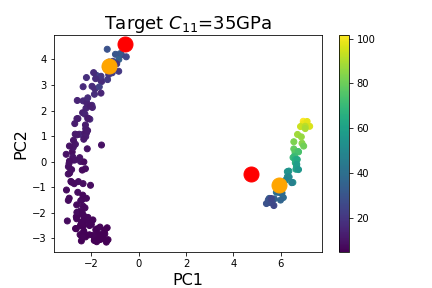}
	\label{fig:optpca}}\\[-0.5ex]
	\subfloat[{}]{
	\includegraphics[scale=0.25,clip, trim=0.5cm 1.2cm 0 0.2]{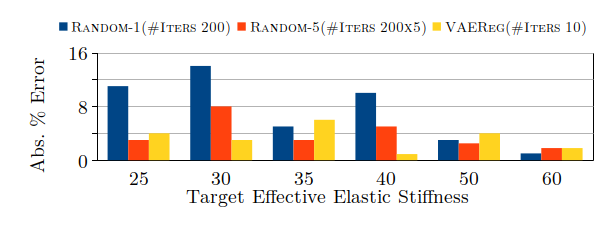}
	\label{fig:invinfoptbarchart}}
	\end{minipage}
	\caption{Optimization based inverse inference (a) For target $C_{11} = 35 GPa$, real microstructures (top row), those obtained through optimization by starting at 5 random initial points of which two were distinct (middle row), and at the points inferred by our method (last row). (b) Visualization in latent space. Optimization starting from the points inferred by our method (red) converges to the orange points in 10 iterations, leading to high-quality solutions (c) The absolute \% error for a range of target $C_{11}$ values. When starting at random points, at least 5 searches, 200 iterations each, are needed to achieve mean error $<$10\%. Starting at points inferred by our method, this is achieved in just 10 iterations}
	\label{fig:invinfopt_optpca}
\end{figure}

To demonstrate this, we implemented simulated annealing optimization in the VAE latent space. The objective function is implemented using the forward prediction model referred to as Reg-only in table~\ref{tab:forwardaccuracy}. For each target property value, we performed the optimization starting at multiple random initial points and those inferred by our method. As an example, figure~\ref{fig:invinfopt} shows the results for target $C_{11}=35GPa$. The top row shows two real microstructures with $C_{11} \approxeq 35GPa$, the middle row shows two distinct solutions found from optimization runs starting with 5 random initial points and the last row shows the result of optimization starting with the points inferred by our method. Similar results for other target property values are shown in appendix~\ref{sec:appB}. 
The solutions are validated through FEM simulations as before. The results are shown in fig.~\ref{fig:invinfoptbarchart}. We found that to ensure that the mean error between the target and achieved property values is within 10\%, we had to run at least 5 searches starting from different random initializations, with at least 200 iterations each. Due to this, the average time for a single inverse inference was $\sim$1.5 hours on a Nvidia V100 GPU, using a reduced-order forward model. Doing this with physics-based FEM simulation model would be clearly infeasible. Comparatively, when starting from the points inferred by our method, we could get to similar accuracy (i.e. mean error $<$10\%) within 10 iterations, which took $\sim$2mins on the same hardware.

This can perhaps be explained as follows: The VAE latent space has regions of high and low probabilities, with high probability regions being quite sparse. The optimization algorithm guided only by the forward-prediction model has no knowledge of this distribution and hence spends considerable amount of time exploring low probability regions. In our approach, the conditional prior used for inverse inference is learned jointly with VAE and property prediction which ensures the conditional prior picks up the true distribution of the latent space. 
Figure~\ref{fig:optpca} shows for target $C_{11} = 35GPa$, how the latent space points inferred by our method are just nudged right to improve the error from $\sim$20\% (please see figure~\ref{fig:invinfbarchart}) to $\sim$6\%.
The plot is a visualization of the latent space in 2D using Principal Components Analysis (PCA). The points are colored by the $C_{11}$ value. The two large red circles are the points inferred by VAE-Reg multi-modal prior (component 1 and 2), which are fed as initial points for optimization. The large orange circles are the points found by optimization after 10 iterations. It can be seen that the points are pushed towards the right high-probability regions, leading to microstructures with $C_{11}$ very close to 35GPa.\\ 
\textbf{Extension to multiple properties} Our model can be easily extended for the case when $c$ is a vector, by suitably changing the sizes of the regressor's output and the generator's input layers. To show this, we extended the dataset by estimating the entire elastic stiffness tensor $C$ through FEM simulations  and applied our method for forward and inverse inference involving the vector $C$. These experiments are discussed in appendix~\ref{sec:appC}.\\
\textbf{Discussion and Limitations}
As discussed already, the number of mixture components $K$ in the prior is treated as a hyperparameter presently. A good value of $K$ could be set using Bayesian global optimization, or can be learned using a suitable hyper-prior for $K$. In some cases, materials scientists may provide a heuristic about the number of practically feasible solutions for a target property value based on domain knowledge of the specific material system under consideration. 
\section{Summary and Conclusions}
We have developed a model for forward and inverse structure-property linkages in materials science by combining VAE and regression. For forward prediction, the combined model performs better than separately trained VAE and regression and is comparable to the state-of-the-art regression-only model. Whereas for inverse inference, the candidate microstructures inferred using our model achieve the target properties reasonably accurately and a local optimization search around these candidates using a reduced-order model quickly reaches target accuracy. Thus a detailed exploration in the small optimal region using physics-based simulations or experiments becomes feasible. 
 \bibliographystyle{splncs04}
 \bibliography{refs}
\appendix
\section{Architecture and Training Details}\label{sec:appA}
\subsubsection{Architecture}
Table~\ref{tab:encarchi} shows the architecture of encoder and regressor. The two networks have a common convolution part after which, two branches fork out - one for encoder and the other for regressor. Output of encoder is of size 16 (number of latent dimensions) and output of regressor is the number of properties (1 or 6). They output $\mu$ and $\sigma$ of $q(z|x)$ and $q(c|x)$ respectively, where $x$ is the microstructure, $z$ the latent representation and $c$ is the property. 
\begin{table}
	%\begin{scriptsize}		
	\centering
	\caption{Encoder Architecture}
	\begin{tabular}{c|l|cccc}
		\hline
		& Layer & \#Filters & Activation & Pool/Stride & Output Size \\ 
		\hline		
		\multirow{8}{*}{\rotatebox[origin=c]{90}{Common}} & Input & - & - & - & (51,51,51,1) \\ 
		&Conv1 & 16 & LReLU(0.2) & Max/2 & (25,25,25,16) \\ 
		&Conv2 & 32 & LReLU(0.2) & Max/2 & (12,12,12,32) \\ 
		&Conv3 & 64 & LReLU(0.2) & Max/2 & (6,6,6,64) \\ 
		&Conv4 & 128 & LReLU(0.2) & Max/2 & (3,3,3,128) \\ 
		&Conv5 & 256 & LReLU(0.2) & Max/2 & (1,1,1,256) \\ 
		&Flatten & - & - & - & 256 \\ 		
		& Dense1 & - & ReLU & - & 2048 \\ 
		\hline
		\multirow{4}{*}{\rotatebox{90}{Encoder}} & Dense2 & - & ReLU & - & 1024 \\ 
		& z\_mean & - & - & - & 16 \\ 
		& Dense2\_1 & - & ReLU & - & 1024 \\ 
		& z\_log\_var & - & - & - & 16 \\ 		
		\hline
		\multirow{4}{*}{\rotatebox{90}{Regressor}}&Dense2\_2 & & ReLU &  & 1024\\
		& c\_mean & & - & & 1 or 6\\
		& Dense2\_3 & & ReLU & & 1024\\
		& c\_log\_var & & - & & 1 or 6\\	
		\hline	
	\end{tabular}
	\label{tab:encarchi}	
	%\end{scriptsize}
\end{table}

Table~\ref{tab:decarchi} shows the decoder architecture. 
\begin{table}
	\centering	
	\caption{Decoder Architecture}
	\begin{tabular}{l|cccc}
		\hline
		Layer & \#Filters & Activation & Upsampling & Output Size\\
		\hline
		Input & - & - & - & 16\\
		Dense1 & - & LReLU(0.2) & - & 32\\
		Dense2 & - & LReLU(0.2) & - & 64\\
		Dense3 & - & LReLU(0.2) & - & 13824\\
		Reshape & - & - & - & (6,6,6,64)\\
		DeConv1  & 64 & LReLU(0.2) & Nearest, 2 & (12,12,12,64) \\
		DeConv2 & 32 & LReLU(0.2) & Nearest, 2 & (24,24,24,32)\\
		DeConv3 & 16 & LReLU(0.2) & Nearest, 2 & (48,48,48,16)\\
		DeConv4 & 1 & Sigmoid & - & (51,51,51,1)\\		
		\hline
	\end{tabular}
	\label{tab:decarchi}
\end{table}
The generator takes the $c$ predicted by regressor as input and produces the $\mu$ and $\sigma$ of conditional prior $p(z|c)$ as output. When $p(z|c)$ is a Gaussian mixture prior with $k$ components, the outputs are - component probabilities $\pi$ (softmax layer of size $k$), and $\mu$'s and $\sigma$'s of all the components (two layers of size $latent\_dim*k$). The two variants of the architecture (Gaussian prior and Gaussian mixture prior) are shown in table~\ref{tab:genarchi} and \ref{tab:genarchi2}, respectively.
\begin{table}
	\begin{minipage}{0.5\textwidth}
		\centering	
		\caption{Gaussian prior}
		\begin{tabular}{l|cc}
			\hline
			Layer & Activation & Output Size\\
			\hline
			Input & - & 1/6\\
			pz mean & - & 16\\
			pz log var & - & 1/6\\
			\hline
		\end{tabular}
		\label{tab:genarchi}
	\end{minipage}
	\begin{minipage}{0.5\textwidth}
		\centering
		\caption{Gaussian mixture prior}
		\begin{tabular}{l|cc}
			\hline
			Layer & Activation & Output Size\\
			\hline
			Input & - & 1/6\\
			Dense & Tanh & 8\\
			pz pis & Softmax & $k$\\
			pz means & - & 16*$k$\\
			pz log vars & - & 16*$k$\\	
			\hline
		\end{tabular}	
		\label{tab:genarchi2}
	\end{minipage}
\end{table}
\subsubsection{Training} 
The model is trained end-to-end using using Adam optimizer with learning rate 0.0005, $\beta_1 = 0.75$, $\beta_2 = 0.999$ and batch size 8. Because of 3D style loss the effective batch size increases many fold, so we use a relatively small batch size to avoid exhausting the memory. For number of latent dimensions, we tried the values ${2, 4, 8, 16, 32, 64}$. We found that with 16 latent dimensions we got good reconstructions while keeping the right entropy of posteriors $q(z|x)$ (i.e., neither peaky nor too wide) and good prediction accuracy. Note that if the posteriors are peaky (low entropy), the encoder is learning almost deterministic representations, possibly leading to a non-continuous latent space which is difficult to sample from. On the other hand, if the posteriors are too wide (high entropy), the representations of most inputs will be similar, loosing the discriminative power. 

The loss function contains three terms - the regression loss, the style (reconstruction) loss and the KL-divergence from prior (regularization). Since these terms have different numerical scales, we first normalize them to the same scale (order
of magnitude) by using suitable multipliers for the regression loss (2) and the KL divergence (20). We then searched for weights of regression loss and KL divergence in ${1, 2, 5, 10, 100}$ and ${0.1, 0.2, 1}$ respectively. We got the best results with weight 10 for regression loss and 1 for KL-Divergence. In the next section we explain why it is necessary to choose 2D slices from all three directions to compute the style loss
between.
\subsubsection{Style loss computation}
\begin{figure*}
	\centering
	\includegraphics[scale=0.38]{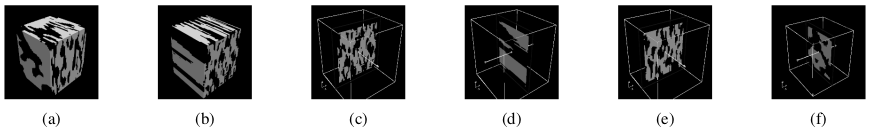}
	\caption{3D Style loss from 2D slices - (a)The observed
		microstructure, (b)Reconstruction by matching 51 slices in x direction, (c)\&(d) Center
		slices of (b) in x and y directions, and (e)\&(f)Center slices of a reconstruction by matching slices in all directions. Features in y direction are left unmatched in (d), as compared to (f).}
	\label{fig:styleloss3d}
\end{figure*}

Style loss between 2 3D inputs is computed using 2D slices so that we can use the pre-trained VGG19. We found that to ensure a match between the 3D features (such as elongation), slices from all three axial directions need to be matched, as shown in figure~\ref{fig:styleloss3d}. Since the input size is 51x51x51, we get 51*3=153 slices. This increases the effective batch size for style loss computation by 153 times, so we use a relatively small batch size of 8.
\section{Additional Experimental Results}\label{sec:appB}
In this section, we present additional experimental results including example reconstructions and inverse inferred microstructures for a range of $C_{11}$ values using our method as well os optimization-based method.
\subsection{Reconstruction}
\begin{figure}[!t]
	\centering	
	\includegraphics[scale=0.22,clip,trim=0 0.5cm 0 0.5]{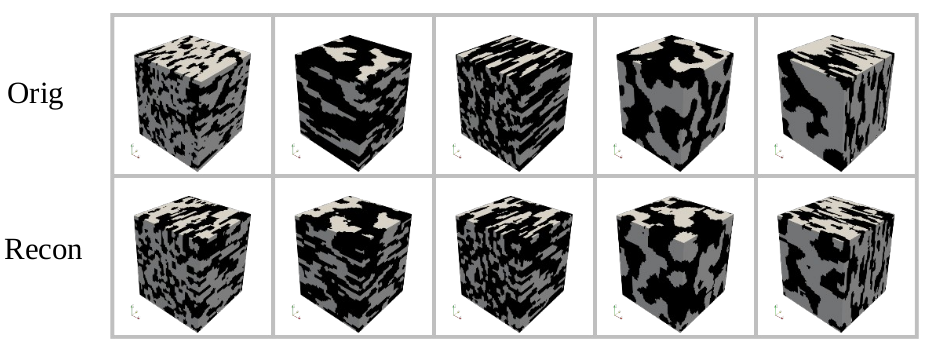}
	\caption{Example reconstructions obtained from the learned VAE model. The inputs and reconstructions are not pixel-wise replicas. Rather, they are statistically equivalent.}
	\label{fig:recons}
\end{figure}

Fig.~\ref{fig:recons} shows example reconstructions of the test set microstructures with different morphological features. The reconstructions are not exactly a replica of the inputs. Rather they are statistically equivalent. The average grain sizes, elongation, and volume fraction of the black phase are similar in the input and the reconstruction. This shows that the latent representation captures these physically significant attributes of the microstructure. 
\subsection{Learned Microstructure Representations}\label{sec:vaereg_latentspace}
\begin{figure}
	\centering
	\subfloat[Vanilla VAE]{
		\includegraphics[scale=0.3]{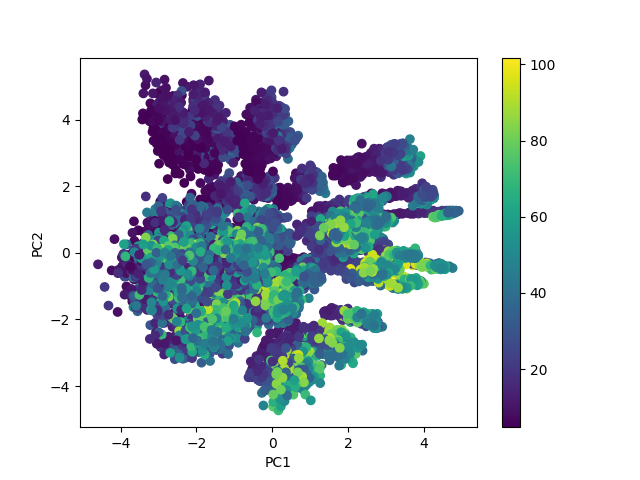}
		\label{fig:latent_space_vae}}
	\subfloat[Combined VAE-Reg]{
		\includegraphics[scale=0.3]{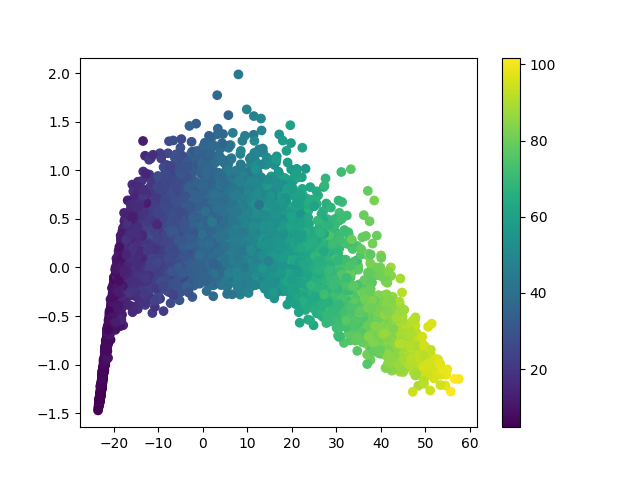}
		\label{fig:latent_space_vaereg}}
	\caption{Comparison of latent spaces learned by vanilla VAE and a combined VAE-regression model}
\end{figure}

The motivation for combining VAE and regression was to enable VAE to learn microstructure features salient for reconstruction and property-prediction. In this section we show that the combined VAE-regression model indeed learns a latent space which is organized by property values and is better suited for linking with properties than a vanilla VAE latent space. Figure~\ref{fig:latent_space_vae} and~\ref{fig:latent_space_vaereg} show the latent spaces learned using a vanilla VAE and a combined VAE-regression model, respectively in 2D using PCA. The points are colored by $C_{11}$ values. The latent space learned by the combined model is clearly better organized for property prediction. Moreover, since the latent space is property-aware, it can facilitate efficient inverse inference (using the conditional prior) avoiding expensive optimization loops. This is explained in detail in the main paper in the section "Comparison with optimization". 
\subsection{Inverse Inference}
\begin{figure}[h]
	\centering	
	\includegraphics[scale=0.25]{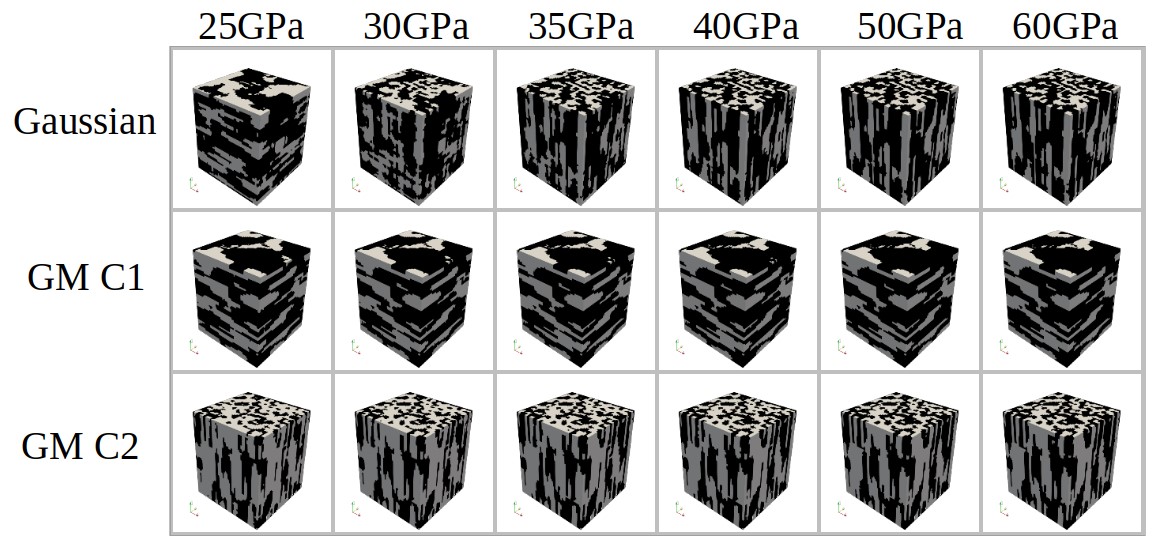}
	\caption{Inverse inference for a range of target $C_{11}$ values using the two priors}
	\label{fig:invinfscalar}
\end{figure}

Figure~\ref{fig:invinfscalar} shows the inverse inferred microstructures for all target values of $C_{11}$ using the two priors. GM C1 and GM C2 denote microstructures obtained from means of the two components of the Gaussian mixture prior, whereas Gaussian denotes those obtained from uni-modal Gaussian prior. While all the microstructures inferred using one component of a Gaussian mixture prior seem to be the same, they actually have increasing volume fractions of black phase for increasing target $C_{11}$ . This is shown in table~\ref{tab:invinfscalarvolfrac}. Please note that the two morphologies are clearly separated by the two components of the mixture prior. Whereas the Gaussian prior tends to infer an average or mix of the two morphologies. This can be seen for the target values $30GPa$ and $35GPa$ which are likely under both the morphologies. It can also be seen from the volume fractions in the inferred microstructures using uni-modal prior shown in table~\ref{tab:invinfscalarvolfrac}.
\begin{table}[h]
	\centering
	\caption{Volume fraction of black phase in the inferred microstructures}	
	\begin{tabular}{c|ccc}				
		\hline
		Target C11 & GM C1 & GM C2 & Gaussian\\
		\hline
		20 & 0.32 & 0.61 & 0.64\\
		25 & 0.35 & 0.63 & 0.63\\
		30 & 0.38 & 0.64 & 0.57\\
		35 & 0.42 & 0.64 & 0.55\\
		40 & 0.44 & 0.65 & 0.56\\
		50 & 0.47 & 0.65 & 0.58\\
		60 & 0.49 & 0.65 & 0.59\\
		80 & 0.5 & 0.66 & 0.6\\
		\hline
	\end{tabular}
	\label{tab:invinfscalarvolfrac}
\end{table}
\subsubsection{Comparison with Optimization}
\begin{figure}
	\centering	
	\subfloat[{$C_{11}=25GPa$}]{
		\includegraphics[scale=0.2]{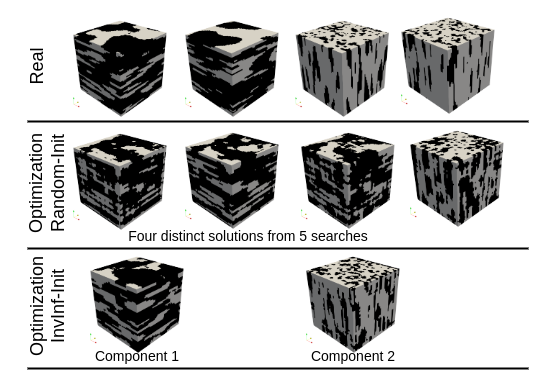}
	}
	\hfil	
	\subfloat[{$C_{11}=35GPa$}]{
		\includegraphics[scale=0.22]{opt_inverse_inference_demo-35}
	}\\	
	\subfloat[$C_{11}=40GPa$]{
		\includegraphics[scale=0.2]{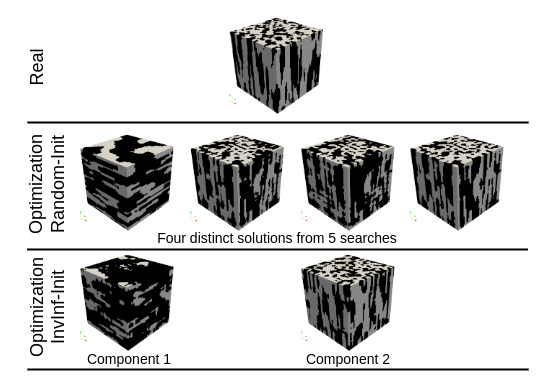}
	}
	\hfil
	\subfloat[$C_{11}=80GPa$]{	
		\includegraphics[scale=0.2]{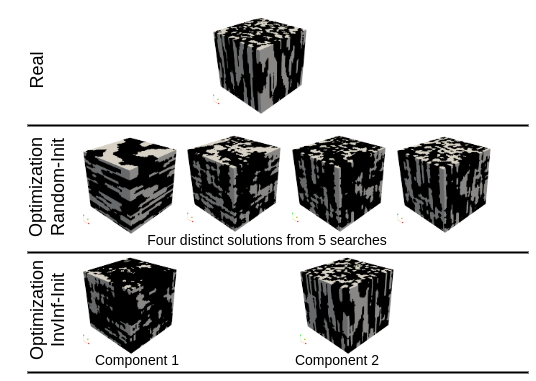}
	}
	\caption{Inverse Inference for various target $C_{11}$ values using Optimization. In each part, top row is real microstructures with the target $C_{11}$, middle row is optimization starting from 5 random initial points and the last row is optimization starting from inverse inference using VAE-Reg.}
	\label{fig:invinfoptextra}
\end{figure}

Figure~\ref{fig:invinfoptextra} shows the results for various target $C_{11}$ values. In each part, the top row shows real microstructures with (approximately) the target $C_{11}$. The middle row shows distinct solutions found from 5 optimization runs (at least 200 iterations each), starting with 5 random initial points and the last row shows result of optimization (10 iterations) starting with the points inferred by the multi-modal prior of VAE-Reg. As can be seen, in each case, starting with the inverse inferred points, optimization quickly reaches diverse optimal solutions, one for each component of the multi-modal prior. 
\section{Extension to multiple properties}\label{sec:appC}
In the experimental results discussed in the main paper, regression variable $c$ was a scalar, $C_{11}$. However, our model can be easily extended for the case when $c$ is a vector, by suitably changing the sizes of regressor's output generator's input layers. To show this, we extended the dataset by estimating other elements of the stiffness tensor through FEM simulations.
\subsection{Extending the dataset} 
Following the FEM simulation method described in~\cite{kalidindidatasource}, we performed additional simulations to estimate all the elements of the stiffness tensor. Elastic stiffness relates stress $\sigma$ with strain $\epsilon$. Since $\sigma$ and $\epsilon$ are second-rank ($3 \times 3$) tensors, a fourth-rank $3 \times 3 \times 3 \times 3$ tensor is required to relate them, i.e.,
$\sigma_{ij} = C_{ijkl} \epsilon_{kl}$ However, due to symmetry of stress and strain, a $6\times6$ matrix is enough to represent $C$. We performed FEM simulations to estimate the complete $C_{6\times6}$ matrix. In the current configurations which lead to orthotropic materials, this matrix is symmetric and the off-diagonal elements after first $3\times3$ sub-matrix are insignificant. 
Thus, we use $C_{11}, C_{21}, C_{31}, C_{22}, C_{32}$ and $C_{33}$, in our experiments.\footnote{please refer to standard texts on generalized Hooke's law, e.g.~\cite{physicsbook}} Note that $C_{11}$ in the new data is the same as $C_{11}$ in the existing data since we are not replacing the original data, rather just augmenting it with additional properties.
\subsection{Forward inference} Table~\ref{tab:forwardaccuracyvector} and \ref{tab:forwardaccuracyvector2} show the forward prediction accuracy using different methods for the diagonal and off-diagonal elements of $C$ respectively. The first block corresponds to separately trained VAE and regression. For regression we experimented with linear regression (LIN), neural network (NN), Gaussian process regression (GPR) and SVR. The second last row corresponds to the regression-only CNN~\cite{kalidindi3dcnn2} model, and the last row corresponds to our VAE-regression model. The numbers are averaged over ten runs with random initialization and train-test splits. 
Note that we extended the state-of-the-art architecture from~\cite{kalidindi3dcnn2} for the vector case by adding more output neurons. Our model consistently performs as good as or better than the state-of-the-art, considering the standard deviations. 
\begin{table*}
	\centering	
	\caption{Forward Prediction Accuracy - Vector Case, Diagonal Elements}	
	\begin{tabular}{l|p{1.5cm}p{1.5cm}|p{1.5cm}p{1.5cm}|p{1.5cm}p{1.5cm}}		
		\hline
		\multirow{2}{*}{Method} & \multicolumn{2}{c}{$C_{11}$} &  \multicolumn{2}{c}{$C_{22}$} &  \multicolumn{2}{c}{$C_{33}$} \\
		\cline{2-7}
		& MAPE & R$^2$ & MAPE & R$^2$ & MAPE & R$^2$ \\
		\hline
		VAE+LIN & 14.25 ($\pm$0.19) & 0.9228 ($\pm$0.0028) & 14.65 ($\pm$0.37) & 0.9138 ($\pm$0.0033) & 12.87 ($\pm$0.26) & 0.9322 ($\pm$0.0019)\\
		VAE+NN & 7.60 ($\pm$0.11) & 0.9717 ($\pm$0.0012) & 7.46 ($\pm$0.20) & 0.9716 ($\pm$0.0014) & 7.259 ($\pm$0.09) & 0.9736 ($\pm$0.0008)\\
		VAE+GPR & 7.23 ($\pm$0.15) & 0.9729 ($\pm$0.0015) & 6.97 ($\pm$0.13) & 0.9737 ($\pm$0.0009) & 6.84 ($\pm$0.11) & 0.9752 ($\pm$0.0013)\\
		VAE+SVR & 6.76 ($\pm$0.12) & 0.9763 ($\pm$0.0013) & 6.56 ($\pm$0.11) & 0.9764 ($\pm$0.0010) & 6.43 ($\pm$0.07) & 0.9779 ($\pm$0.0010)\\
		\hline
		Reg-Only (\cite{kalidindi3dcnn2}) & 4.27 ($\pm$0.29) & 0.9923 ($\pm$0.0009) & 4.16 ($\pm$0.29) & 0.9921 ($\pm$0.0010) & 4.15 ($\pm$0.3) & 0.9923 ($\pm$0.0011)\\
		VAE-Reg (ours) & \textbf{3.99} ($\pm$0.3) & \textbf{0.9932} ($\pm$0.0010) & \textbf{3.77} ($\pm$0.18) & \textbf{0.9933} ($\pm$0.0010) & \textbf{3.82} ($\pm$0.16) & \textbf{0.9934} ($\pm$0.0006)\\
		\hline
	\end{tabular}
	\label{tab:forwardaccuracyvector}
\end{table*}
\begin{table*}
	\centering	
	\caption{Forward Prediction Accuracy - Vector Case, Off-diagonal Elements}
	\begin{tabular}{l|p{1.5cm}p{1.5cm}|p{1.5cm}p{1.5cm}|p{1.5cm}p{1.5cm}}		
		\hline
		\multirow{2}{*}{Method} & \multicolumn{2}{c}{$C_{21}$} &  \multicolumn{2}{c}{$C_{31}$} &  \multicolumn{2}{c}{$C_{32}$} \\
		\cline{2-7}
		& MAPE & R$^2$ & MAPE & R$^2$ & MAPE & R$^2$ \\
		\hline
		VAE+LIN & 19.4476 ($\pm$0.3370) & 0.8657 ($\pm$0.0030) & 18.9616 ($\pm$0.2750) & 0.8734 ($\pm$0.0034) & 17.0849 ($\pm$0.2472) & 0.894 ($\pm$0.0020)\\
		VAE+NN & 9.6434 ($\pm$0.3462) & 0.9575 ($\pm$0.0031) & 9.2863 ($\pm$0.1511) & 0.961 ($\pm$0.0013) & 9.2004 ($\pm$0.2123) & 0.9608 ($\pm$0.0017)\\
		VAE+GPR & 8.1671 ($\pm$0.1726) & 0.9644 ($\pm$0.0016) & 7.9095 ($\pm$0.1849) & 0.9676 ($\pm$0.0020) & 8.0401 ($\pm$0.1558) & 0.9662 ($\pm$0.0014)\\
		VAE+SVR & 7.59 ($\pm$0.15) & 0.969 ($\pm$0.0014) & 7.36 ($\pm$0.17) & 0.9717 ($\pm$0.0023) & 7.58 ($\pm$0.12) & 0.9695 ($\pm$0.0014)\\
		\hline
		Reg-Only (\cite{kalidindi3dcnn2}) & 5.82 ($\pm$0.43) & 0.9855 ($\pm$0.0021) & 5.78 ($\pm$0.67) & 0.9858 ($\pm$0.0031) & 6.0 ($\pm$0.72) & 0.9846 ($\pm$0.0031)\\
		VAE-Reg (ours) & \textbf{5.48} ($\pm$0.32) & \textbf{0.9872} ($\pm$0.0011) & \textbf{5.33} ($\pm$0.21) & \textbf{0.9880} ($\pm$0.0009) & \textbf{5.35} ($\pm$0.23) & \textbf{0.9880} ($\pm$0.0018)\\		
		\hline
	\end{tabular}	
	\label{tab:forwardaccuracyvector2}
\end{table*}
\subsection{Inverse Inference}
\begin{figure}[!t]
	\centering	
	\includegraphics[scale=0.28]{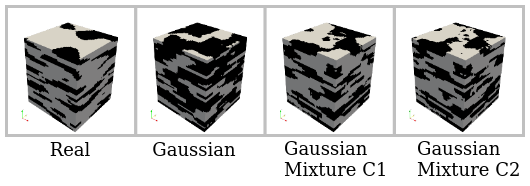}
	\caption{Inverse inference for a vector of properties using Gaussian and Gaussian mixture priors.} 
	\label{fig:invinfvector}
\end{figure}

We perform inverse inference for vectors $[C_{11},\ldots, C_{33}]$ chosen from the test data since it's easier to validate the results on them as reference structures with these properties are already available in the test data. We discuss the inverse inference for an example property vector: 
\begin{multline*}
[C_{11}~~C_{22}~~C_{33}~~C_{21}~~C_{31}~~C_{32}] \\= 
[8.24~~38.07~~43.22~~4.44~~3.99~~10.70]
\end{multline*}
Fig.~\ref{fig:invinfvector} shows the observed and inferred microstructures for this target elastic stiffness vector. The leftmost image is the observed microstructure, while the rest (from the second) are inferred using the mean of the uni-modal Gaussian prior, and the two means of the  Gaussian mixture prior with two components. The microstructures inferred from the mixture prior are closer to the real microstructure and hence achieve the target set of properties better (as explained next). 

\begin{figure}{h}
	\centering	
	\includegraphics[scale=0.25]{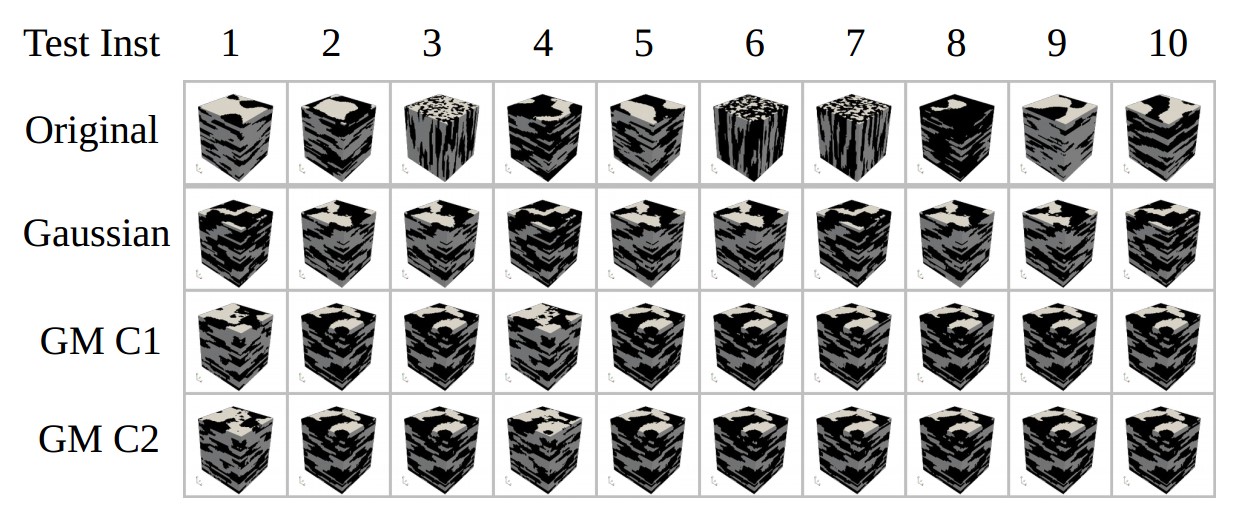}
	\caption{Inverse inference for target vectors $[C_{11} \ldots C_{33}]$ shown in table 8}
	\label{fig:invinfvectorsupp}
\end{figure}

Inverse inference was performed similarly for 10 target property vectors $C_{11} \ldots C_{33}$ from the test-set. Table~\ref{tab:invinfvectortargets} shows the property vectors and figure~\ref{fig:invinfvectorsupp} shows microstructures inferred for these target vectors. Note that many of the target vectors are quite similar and so the inferred microstructures too are similar. 
\begin{table}
	\centering
	\begin{scriptsize}
		\caption{Target vectors $[C_{11} \ldots C_{33}]$ from the test data}		
		\begin{tabular}{c|p{1cm}p{1cm}p{1cm}p{1cm}p{1cm}p{1cm}}				
			\hline
			No. & $C_{11}$ & $C_{21}$ & $C_{31}$ & $C_{22}$ & $C_{32}$ & $C_{33}$\\
			\hline
			1 & 8.24 & 4.44 & 3.99 & 38.07 & 10.7 & 43.22\\
			2 & 14.64 & 6.94 & 7.29 & 64.97 & 19.02 & 63.03\\
			3 & 14.07 & 7.09 & 6.99 & 67.95 & 19.72 & 65.22\\
			4 & 7.83 & 3.11 & 3.64 & 31.74 & 8.84 & 40.6\\
			5 & 16.59 & 7.28 & 7.45 & 66.93 & 19.65 & 65.84\\
			6 & 12.75 & 6.42 & 6.39 & 74.23 & 21.55 & 70.61\\
			7 & 9.8 & 4.68 & 4.57 & 51.51 & 15.23 & 57.91\\
			8 & 16.83 & 7.75 & 7.51 & 80.56 & 24.3 & 79.55\\
			9 & 16.83 & 8.07 & 7.9 & 79.87 & 24.4 & 79.67\\
			10 & 10.42 & 5.02 & 4.84 & 53.46 & 14.83 & 54.41\\
			\hline
		\end{tabular}
		\label{tab:invinfvectortargets}		
	\end{scriptsize}
\end{table}

The inferred microstructures were then validated through FEM simulations to get the achieved properties, as before. Table~\ref{tab:invinfvectoracc} shows the mean absolute percentage error between target and achieved values of $C_{11},\ldots,C_{33}$ for the two priors. The solutions inferred using Gaussian mixture prior consistently achieve the target sets of properties with better accuracy.
\begin{table}
	\centering
	\caption{Abs. \% Error Between Target and Achieved $C_{ij}$}
	\begin{tabular}{l|c|c|c|c|c|c}
		\hline
		\multirow{2}{*}{Prior} & \multicolumn{6}{c}{Mean Absolute \% Error}\\
		\cline{2-7}
		& $C_{11}$ & $C_{21}$ & $C_{31}$ & $C_{22}$ & $C_{32}$ & $C_{33}$ \\
		\hline
		Gaussian & 35 & 38 & 36 & 29 & 34 & 27 \\
		Gaussian Mixture & 13 & 13 & 12 & 12 & 13 & 9 \\
		\hline
	\end{tabular}
	\label{tab:invinfvectoracc}
\end{table}

\end{document}